\documentclass{article}

\usepackage[preprint]{neurips_2021}

\usepackage[utf8]{inputenc} % allow utf-8 input
\usepackage[T1]{fontenc}    % use 8-bit T1 fonts
\usepackage{hyperref}       % hyperlinks
\usepackage{url}            % simple URL typesetting
\usepackage{booktabs}       % professional-quality tables
\usepackage{amsfonts}       % blackboard math symbols
\usepackage{nicefrac}       % compact symbols for 1/2, etc.
\usepackage{microtype}      % microtypography
\usepackage{colortbl}
\usepackage[dvipsnames]{xcolor}
\usepackage{graphicx}
\usepackage{amsmath} %maths symbols 1
\usepackage{amssymb} % maths symbols 2
\usepackage{amsfonts} % maths symbols 3
\usepackage{multirow} % added for table 1
\usepackage{algorithm} % algorithm1
\usepackage{algorithmic} % algorithmic
\usepackage{stmaryrd}
\usepackage{bbm}
\usepackage{mathtools, nccmath}% to deal with equations
\usepackage{colortbl}
\usepackage{setspace} % spacing for equations
\usepackage{enumitem}
\usepackage{booktabs} % for professional tables
\usepackage{pifont}% http://ctan.org/pkg/pifont

\newcommand{\cmark}{\textcolor{OliveGreen}{\ding{51}}}%
\newcommand{\xmark}{\textcolor{red}{\ding{55}}}%

\definecolor{Gray}{gray}{0.85}

\newcommand{\ours}{RED\,}
\newcommand{\oursvir}{RED}

\title{\ours: Looking for Redundancies for Data-Free Structured Compression of Deep Neural Networks}

\author{%
    Edouard Yvinec\\
  Datakalab \& ISIR\\
  114 boulevard Malesherbes\\
  75017 Paris, France \\
  \texttt{ey@datakalab.com} \\
   \And
   Arnaud Dapogny \\
   Datakalab \\
   114 boulevard Malesherbes\\
   75017 Paris, France \\
   \texttt{ad@datakalab.com} \\
   \And
   Matthieu Cord \\
   Lip6 \\
   4, place Jussieu 75005 Paris, France \\
   \texttt{matthieu.cord@lip6.fr} \\
   \And
   Kevin Bailly \\
   ISIR \\
   4 Place Jussieu 75005 Paris, France \\
   \texttt{bailly@isir.upmc.fr} \\
}

\begin{document}

\maketitle

\begin{abstract}
  Deep Neural Networks (DNNs) are ubiquitous in today's computer vision landscape, despite involving considerable computational costs. The mainstream approaches for runtime acceleration consist in pruning connections (\textit{unstructured} pruning) or, better, filters (\textit{structured} pruning), both often requiring data to retrain the model.
  In this paper, we present \oursvir, a data-free structured, unified approach to tackle structured pruning. First, we propose a novel adaptive hashing of the scalar DNN weight distribution densities to increase the number of identical neurons represented by their weight vectors. Second, we prune the network by merging redundant neurons based on their relative similarities, as defined by their distance. Third, we propose a novel uneven depthwise separation technique to further prune convolutional layers. We demonstrate through a large variety of benchmarks that \ours largely outperforms other data-free pruning methods, often reaching performance similar to unconstrained, data-driven methods.
\end{abstract}
\section{Introduction}\label{sec:introduction}
% \begin{figure}[t]
% \begin{center}
% \includegraphics[width = 0.5\linewidth]{new_main.pdf}
% \caption{\ours overview: \textbf{(a)}, adaptive hashing dramatically restricts the number of neuron weight values, (color-maps on the right). \textbf{(b)}, Redundant neurons are merged to prune the network with minimal output alteration. \textbf{(c)}, convolutional layers are factored in uneven depthwise convolutions. This data-free procedure allows for a significant memory footprint and runtime reduction.}
% \label{fig:main}
% \end{center}
% \vskip -0.2in
% \end{figure}\noindent
Modern Deep Neural Networks (DNNs) have become the mainstream approach in machine learning in general and in computer vision in particular, with CNNs achieving outstanding performance on various tasks such as object classification \citep{he2016deep}, detection \citep{he2017mask} or segmentation \citep{chen2017deeplab}. However, DNNs usually reach high requirements in terms of computational runtime. This prevents most state-of-the-art models to be deployed, most notably on edge devices. To address this shortcoming, a number of approaches for DNN compression have been proposed over the past few years.
% On the one hand, reducing the memory footprint of DNNs is usually addressed via quantization \cite{stock2019and} or hashing \cite{han2015deep} techniques. On the other hand, 
Architecture compression constitutes a convenient and popular way to address this runtime limitation, involving pruning as well as tensor decomposition techniques \citep{cheng2017survey}. It consists in either removing connections, \textit{i.e.} an \textit{unstructured} way or suppressing or reordering specific channels or filters \textit{i.e.} a \textit{structured} fashion. Although the former usually removes more weights than the latter \citep{park2020lookahead}, unstructured compression has the drawback to produce sparse weight matrices, which require dedicated hardware or libraries \citep{han2016eie} for real-case runtime improvements. Furthermore, these methods can also be divided in \textit{data-driven} vs. \textit{data-free} methods. While data-free methods are far more convenient for privacy concerns, as some data may be confidential (e.g. health data or military), they are still significantly outperformed by data-driven methods. 
Hence, despite recent work \citep{kim2020neuron, tanaka2020pruning}, data-free architecture compression remains a challenging and promising domain with room for improvements.
In this paper, we propose \ours, a novel data-free structured compression framework. First, \ours leverages a novel adaptive scalar hashing of the layer-wise weight distributions to introduce redundancies in DNNs. In particular, we show that this hashing allows to introduce vector redundancies (\textit{i.e.} neurons that perform the same operation) as well as tensor redundancies (\textit{i.e.} low-rank flattened convolution kernels). These redundancies can respectively be exploited by applying similarity-based neuron merging, as well as a novel uneven depthwise separation of convolutional layers. To sum it up, our contributions are:
%
% In this paper, we empirically find that value-wise and structural REDundancies among weight values can emerge from hashing the weight distributions.
% Based upon this observation we propose \oursvir, a novel data-free structured compression method.
% \ours consists in 3 steps: First, we propose a novel adaptive, data-free hashing algorithm to reduce the number of different weight values. 
% Second, stemming from the fact that the former step introduces redundancies among neurons, we apply the neuron merging algorithm from \cite{srinivas2015data}. 
% Third, we factor convolutional layers in uneven depthwise separable convolutions which is an improvement upon \cite{guo2018network}. To sum it up, our contributions are:
\begin{itemize}
\item An adaptive scalar weight hashing technique based on local extrema search of the weight distribution density, that introduces redundancies among neurons without significantly altering the predictive function.
\item A method for exploiting redundancies at the vector level with similarity-based neuron merging, and at the tensor level, with an uneven depthwise separation of convolutional layers that factors spatially redundant components.
\item We introduce \oursvir, a portable method for data-free structured DNN compression that significantly outperforms state-of-the-art data-free methods and often rivals existing data-driven approaches.
\end{itemize}
% The rest of the paper is organized as follows: in Section \ref{sec:related} we review recent state-of-the-art methods for DNN compression. In Section \ref{sec:methodology}, we provide an overview of \ours for data-free structured DNN compression. In Section \ref{sec:experiments} we quantitatively validate \ours both for memory footprint and runtime reduction on a variety of benchmarks before providing concluding remarks in Section \ref{sec:conclusion}.

\section{Related Work}\label{sec:related}
% \subsection{Compressing Networks \textit{via} Hashing}
\label{sec:related_hashing}
% Working around memory constraints has previously been explored in the context of deep learning. Most algorithm are extensions or variants of k-means (\cite{lloyd1982least}) which requires a prior on the studied distribution to determine a fitting value for $k$.
% Recent approaches clusters weights values \cite{chen2015compressing} prior to applying a quantization scheme \cite{krishnamoorthi2018quantizing} to reduce the bit-wise representations. 
% This clustering step is usually formulated as an optimisation problem involving the DNN layer activations over the training database \cite{wang2019haq,stock2019and}, making these methods data-driven. Furthermore, the size of the codebook is an hyperparameter that has to be set beforehand for each layer \cite{wang2019haq,stock2019and}, making these methods non-adaptive w.r.t. the weight distribution. Approaches such as \cite{liu2017deep} rely on estimating an optimal codebook size using labelled data but these techniques have not yet been exploited for DNN weights hashing. By contrast, we propose a weight distribution density-adaptive and fully data-free hashing method.
% \subsection{Compressing Networks \textit{via} Pruning}
\label{sec:related_pruning}

Most architecture compression methods rely on an underlying approximation of the predictive function to later perform pruning, wether it can is \textit{unstructured} or \textit{structured} (as stated in \cite{renda2020comparing}), \textit{data-driven} or \textit{data-free}, \textit{magnitude-based} or \textit{similarity-based}.

\paragraph{Predictive function approximation:}

%It can come from the removal of specific connections in magnitude-based pruning \citep{frankle2018lottery,lin2020dynamic,park2020lookahead,lee2019signal} or similarity-based pruning \citep{srinivas2015data,kim2020neuron}. In this work we distinguish the pruning aspect from the approximation made on the predictive function.

Perhaps one of the most studied such approximation is quantization \citep{nagel2019data, meller2019same, zhao2019improving}. Quantization consists in mapping DNN weights to a finite, regular grid of values. It generally aims at reducing the inference time by coding this restricted set of values with fewer bytes (e.g. float16, int8 quantization), although it is generally non-adaptive to the weight distribution. Hashing constitutes another intuitive approach for predictive function approximation. Most hashing algorithms are extensions or variants of k-means \citep{lloyd1982least} which requires a prior on the studied distribution to determine a fitting value for the number of clusters. This step can also be formulated as a learning problem \citep{wang2019haq,stock2019and}, but requires data as a consequence.
In this work we propose a data-free, prior-free and adaptive hashing of the scalar weight distribution, which introduces redundancies in DNNs.

\paragraph{Structured Pruning:}
On the one hand, unstructured approaches \citep{frankle2018lottery,lin2020dynamic,park2020lookahead,lee2019signal} consist in removing individual weights: hence, these methods rely on sparse matrices for implementation, which require dedicated hardware to fully exploit the reduction at inference time. On the other hand, the so-called structured approaches \citep{liebenwein2019provable, li2016pruning, he2018soft, luo2017thinet} aim at removing specific filters, channels or neurons. Although the latter usually results in less impressive raw pruning ratios as compared to the former, they allow significant runtime reduction without using dedicated hardware.
\paragraph{Data-free Pruning:}
Most pruning methods can be classified as \textit{data-driven} as they involve, to some extent, the use of a training database. \citet{lee2019signal} uses the drift of DNN weights from their initial values during training to select and replace irrelevant weights.  The Hrank method \citep{lin2020hrank} consists in removing low-rank feature maps, based on the observation that the latter usually contain less relevant information. \citet{lin2020dynamic} extend the single layer magnitude-based weight pruning to a simultaneous multi-layer optimization, in order to better preserve the representation power during training. Other approaches, such as \citep{liebenwein2019provable, meng2020pruning}, train an over-parameterized model and apply an absolute magnitude-based pruning scheme which removes a number of channels or neurons but generally causes accuracy drop. To address this problem, most of these methods usually fine-tune this pruned model for enhanced performance \citep{liu2018rethinking,gale2019state,frankle2018lottery}. Nevertheless, there exists a number of so-called \textit{data-free} approaches which do not require any data or fine-tuning of the pruned network, however usually resulting in lower pruning ratios. For instance, \citet{tanaka2020pruning} is a data-free pruning method with lower performances but still addresses the layer-collapse issue (where all the weights in a layer are set to $0$) by preserving the total synaptic saliency scores.
\paragraph{Similarity-based Pruning:}
All the aforementioned methods remove connections during training \textit{via} more or less adaptive or learnable thresholds under which DNN weights are pruned. Hence, such paradigm constitutes an \textit{absolute} pruning heuristic. However, \textit{relative} methods based on comparison between neurons or feature maps have also been proposed in the literature. For instance, \citet{ayinde2019redundant} use a graph-based-group-average technique to define proximity between feature maps in order to prune filters. 
Recently, in \cite{kim2020neuron}, similarity-based data-free pruning is applied by merging neurons. The authors decompose the weight tensor into new weights and a scaling matrix to fold into the next layer. However, in this approach, merging is performed at the tensor-level, rather than using a fine-grained pairwise similarity. \cite{srinivas2015data}, conversely, propose to merge together neurons of each layer based on pairwise vector similarity between them. However, their approach doesn't exploit scalar weight approximation, nor does it take into account tensor-level redundancies, resulting in low pruning ratios.
\newline\noindent
In this work, we introduce redundancies in DNNs by performing scalar hashing of the layer-wise weight distributions. In particular, we show that this hashing allows to introduce vector redundancies (\textit{i.e.} neurons that perform the same operation) as well as tensor redundancies (\textit{i.e.} low-rank flattened convolution kernels). These redundancies can respectively be exploited by applying similarity-based neuron merging in the same vein as \cite{srinivas2015data}, as well as a novel uneven depthwise separation of convolutional layers.

\section{Introducing Redundancies via Adaptive Weights Hashing}\label{sec:hashing}
\begin{figure}[t]
\begin{center}
\includegraphics[width = 0.9\linewidth]{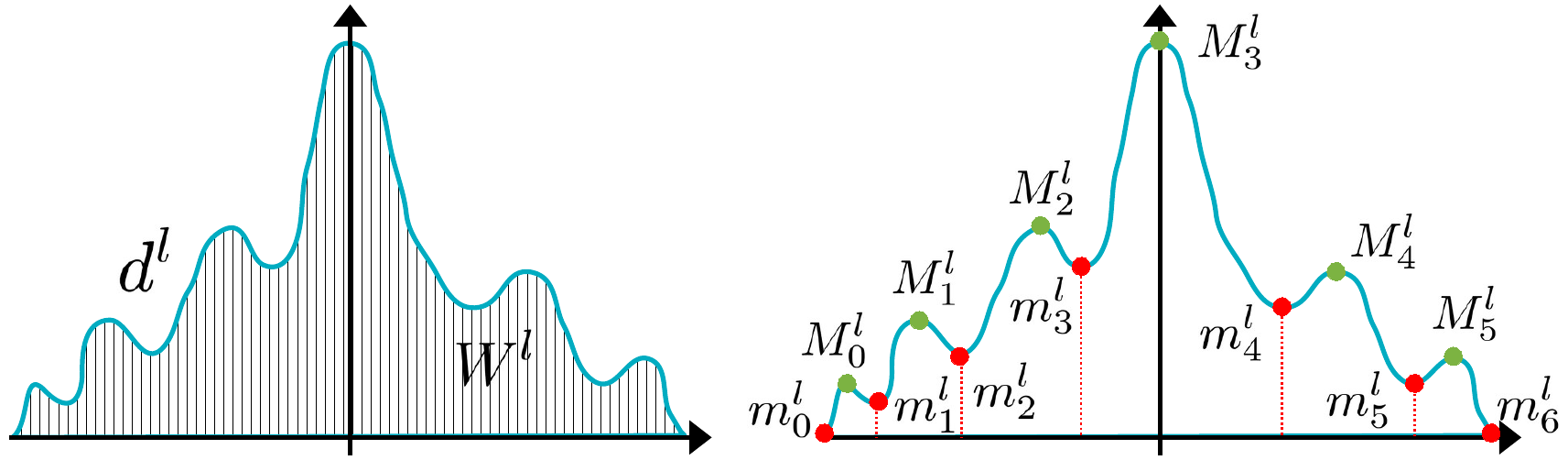}
\vskip -0.1in
\caption{Illustration of the proposed adaptive scalar weight hashing. First (left), we find the estimate of the density function $d^l$ associated to the weights values $W^l$. Second (right), we find the local extrema ${(m_k^l)}_{k \in K^-_l}$ and ${(M_k^l)}_{k \in K^+_l}$ of $d^l$. Then we assign the new values $\tilde W^l \in {K_l^+}^{n_l \times n_{l-1}}$.}% This figure only serves the purpose of clarity (for real distributions see Figure \ref{fig:kde}).}
\label{fig:hashing}
\end{center}
\vskip -0.2in
\end{figure}
Let's consider a DNN $f$ with $L$ layers $f = f_L \circ \cdots \circ f_1$. 
Each layer indexed by $l$ is defined by parameters $W^l$ for a $n_l$-dimensional output.
As DNN weights take values in $\mathbb{R}$, the probability for two values to be equal is almost surely zero: as a consequence, the probability to have redundancies in DNNs (e.g. two neurons that perform the same operation, \textit{i.e.} that have the same weight vector) is also zero, thus limiting the possible simplification of the predictive function $f$.
To deal with this, we propose to simplify $f$ by first hashing the scalar weight distribution. As illustrated on Figure \ref{fig:hashing}, we approximate the density function of the weights distribution for a each layer $f_l$ using Kernel Density Estimation (KDE):

\begin{equation}
    d^l : \omega \mapsto \frac{1}{n_l \times n_{l-1} \Delta_l}\sum_{w \in W^l} K\left( \frac{\omega-w}{\Delta_l} \right)
\end{equation}

where $K$ is the density of a Gaussian kernel with bandwidth $\Delta_l$. 
Then, we estimate the local minima $K^-_l = {(m_k^l)}_{k}$ and maxima $K^+_l = {(M_k^l)}_{k}$ of $d_l$ by computing its values over a discreet grid with range $[\min\{w \in W^l\}; \max\{w \in W^l\}]$. 
Because the KDE provides a continuous density function, the intermediate value theorem guarantees that $|K^+_l| + 1 = |K^-_l|$. 
We can thus partition $\mathbb{R}$ in $|K^+_l|$ intervals with boundaries defined by the local minima and assign the value of the local maximum to all parameters within the corresponding intervals.
Assuming the ${(m_k^l)}_{k}$ and ${(M_k^l)}_{k}$ sorted, with $m_0^l = - \infty$ and $m_{|K^-_l|-1}^l = +\infty $, for every weight $w$ in $W^l$ there exists $k$ such that $w \in \left[m_k^l ; m_{k+1}^l \right[ \quad \text{ and }\quad \tilde w = M_k^l$. This defines the hashed layer $\tilde f^l$ with weights $\tilde W^l$. Note that this method is adaptive and has no prior on the weight distribution contrary to k-means.
\newline\noindent
In practice, we find that DNN weights concentrate around a limited number of local modes: hence, the proposed adaptive hashing dramatically reduces the number of different values that the weights can take, and introduces redundancies both at the vector and tensor level.
Optionally, to conveniently enable further compression, we define (global and per-layer) contrast hyperparameters $\tau=\frac{1}{L}\sum_{l=1}^L \tau^l$. 
These hyperparameters allow to define a distance threshold (relative to each layer's weight values range) under which two modes are collapsed to the dominant one, further increasing redundancies. We show in the experiments that hashing with $\tau=0$ do not significantly alter the predictive function, thus setting $\tau>0$ allows to find suitable trade-off between runtime acceleration and accuracy. 

%This hashing needs to satisfy a number of practical constraints: First, it has to be adaptive which eliminates methods such as k-means which require a prior on the distribution in order to select $k$. Second, the method needs to be data-free which eliminates learning-based approaches. For these reasons we propose a novel adaptive data-free scalar hashing using kernel density estimation (KDE).

\section{Exploiting Redundancies in DNNs}\label{sec:merge+depthwise}
\begin{figure}[t]
\begin{center}
\includegraphics[width = 0.85\linewidth]{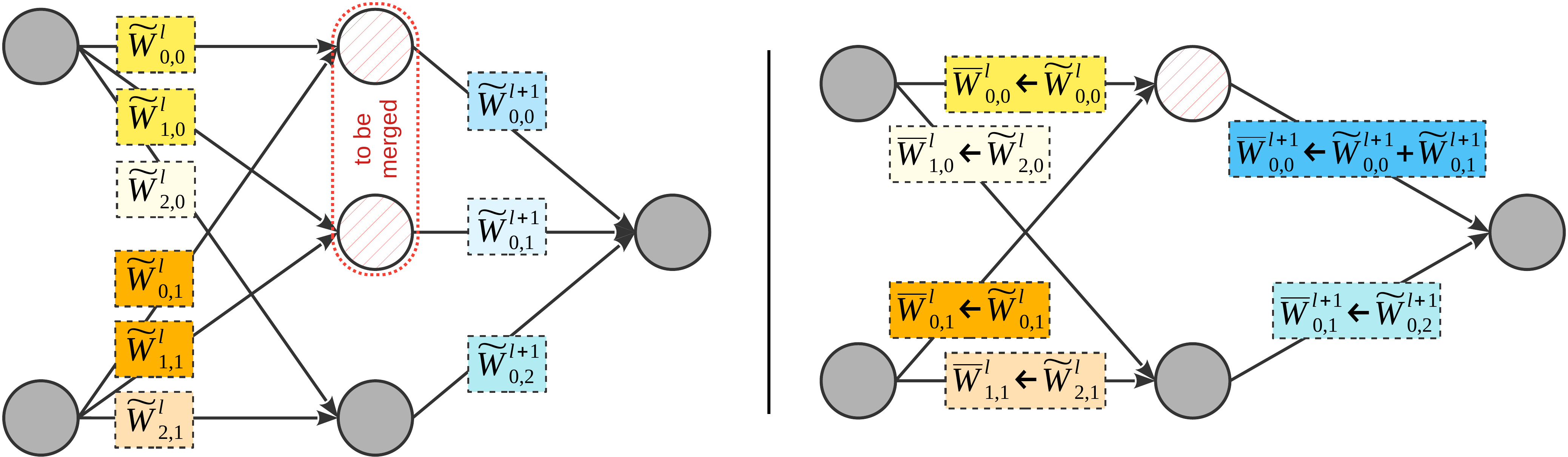}
\vskip -0.1in
\caption{Neuron merging in the case of a fully-connected layer $l$ with weights $\tilde W^l_{i,j}$. Similar colors indicate equal weight values, e.g. $\tilde W^l_{0,0} = \tilde W^l_{1,0}$. The pruned network weights $\bar W$ are obtained by merging the first 2 neurons of layer $l$ and simply summing the corresponding weights in layer $l+1$.}
\label{fig:pruning}
\end{center}
\vskip -0.2in
\end{figure}

As the hashed weights $\tilde W^l$ take values in a finite set, redundancies are much more likely to occur both at the vector and tensor level.
\paragraph{Vector Redundancies:}
First, let's consider the case of a two-layers fully-connected neural network $f$ with an element-wise activation function $\sigma$ and no biases, \textit{i.e.} $ \tilde f :  z \mapsto \tilde W^{2} \sigma(\tilde W^1 z)$ with $\tilde W^1 \in {K^+_1}^{n_1 \times n_0}$ and $\tilde W^2 \in {K^+_2}^{n_2 \times n_1}$ the hashed parameters. 
Let's also assume that we have $\bar n_1 < n_1$ distinct neurons, \textit{i.e.} $\bar n_1$ distinct rows in $\tilde W^1$.  
Let $ \bar W^1$ be the sub-matrix  of $\tilde W^1$ containing all the distinct rows of $\tilde W^1$ only once and $\bar W^2$ the matrix such that all columns from $\tilde W_2$ that were applied to identical neurons of $\tilde W^1$ are summed. 
Then, for each output dimension $i$ we have:
\begin{equation}
    \Big(\tilde f (z)\Big)_{i} = \sum_j^{n_1} \tilde W^2_{i,j} \sigma\left( \sum_k^{n_1} \tilde W^1_{j,k} z_k \right) = \sum_j^{\bar n_1} \bar W^2_{i,j} \sigma\left( \sum_k^{\bar n_1} \bar W^1_{j,k} z_k \right) = \Big(\bar f (z)\Big)_{i}
\end{equation}
In other words, we can merge identical neurons (rows of $\tilde W^1$) and sum the corresponding weights in $\tilde W^2$ without altering the (hashed) layers outputs, as illustrated in Figure \ref{fig:pruning} on a simple case.
This process can straightforwardly extended to neural networks with $L$ layers by repeating this process from the first to last layer. Furthermore, it can be adapted to:
\begin{itemize}
    \item Layers with bias by considering $\tilde W' = (\tilde W \quad \tilde b)$ and $z' = (z^T \quad 1)^T$
    \item Convolutional layers, by using a rewriting of the kernel, following \cite{ma2017equivalence}
    \item Batch-Normalization layers, by folding them like in \cite{nagel2019data}
    \item Skip Connections, where each output channel is computed using the corresponding weights in $\tilde W^l$ and $\tilde W^{l+k}$. Merge is performed on the two added layers simultaneously by considering the concatenation $\begin{pmatrix}\tilde W^l & \tilde W^{l+k}\end{pmatrix}$.
\end{itemize}
In the case of depthwise convolutions \citep{sandler2018mobilenetv2,howard2019searching}, each filter is applied to a distinct input. As a consequence, two filters intrinsically cannot perform identical operations and thus cannot be merged.\newline\noindent
% Note that the merging can't be performed on depthwise convolutions (\cite{sandler2018mobilenetv2}, \cite{howard2019searching}). 
% Also note that in the case of skip connections (\cite{he2016deep}) we simply see the two added layers as one layer where each output map is defined by the corresponding weights in each added layers.\newline\noindent
In practice, we find that after hashing, such exact neuron merging already allows to remove significant number of parameters. However, to obtain larger pruning ratios, this merging can be relaxed: for a layer $l$, we sort the set of pairwise distances between neurons or filters (e.g. as defined by the Euclidean distance between their weight vectors) and merge the $\alpha^l\%$ closest neurons by taking their average weight values. We thus define (global and layer-wise) hyperparameters $\alpha=\frac{1}{L}\sum_{l=1}^L \alpha^l$.

\begin{figure}[t]
\begin{center}
\includegraphics[width = 0.75\linewidth]{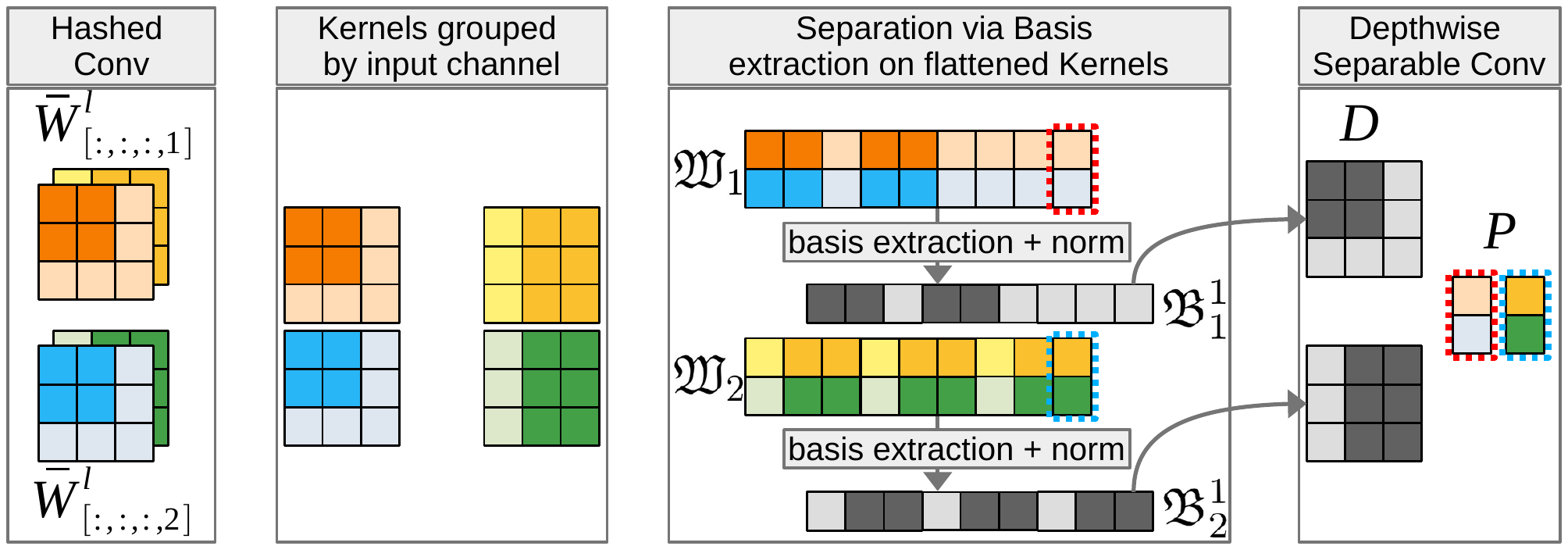}
\vskip -0.1in
\caption{Tensor-level Simplification \textit{via} of our uneven depthwise separation method for a layer with 2 input and 2 output channels. Restrictions of filters to each input channel $i$ are flattened and concatenated along the output channel axis, to form matrices $\mathfrak{W}_i$. If these matrices have rank $r_i$ we extract a basis $\mathfrak{B}_i^j$ (for $j$ in $\llbracket 1;r_i\rrbracket$) of the rows of each $\mathfrak{W}_i$ and normalize it so that the last element is equal to 1. Then we deduce the weights of the pointwise kernels $P$ from the $(\mathfrak{B}_i^j)$ and the $\bar W$.}
\label{fig:depthwise}
\end{center}
\vskip -0.2in
\end{figure}\noindent
\paragraph{Tensor Redundancies:}
The proposed hashing method also induces tensor-level redundancies in regular 2D convolutional layers: to handle these redundancies we propose a novel uneven depthwise separation method. 
Let's consider a kernel $\bar W \in {K^+}^{w\times h \times n_{\text{in}} \times n_{\text{out}}}$ for a layer $l$ after either the hashing or neuron merging steps (as merging and the proposed uneven depthwise separation steps can be applied in any order, see Appendix A.1).
$\bar W$ can be expressed as a depthwise separable convolution if there exists $D \in \mathbb{R}^{w\times h \times n_{\text{in}} \times 1}$ and $P \in \mathbb{R}^{1\times 1 \times n_{\text{in}} \times n_{\text{out}}}$ the respective weights of depthwise and pointwise convolution such that :
\begin{equation}
    \forall y,x,i,j, \quad \bar W_{[y,x,i,j]} = {D}_{[y,x,i,1]} {P}_{[1,1,i,j]}
\end{equation}
This condition can be expressed in terms of matrix ranks, as illustrated on Figure \ref{fig:depthwise}.
We define the matrix $\mathfrak{W}_i$ as the restrictions of filters to input channel $i$, flattened and concatenated along the output channel axis:
\begin{equation}
    \mathfrak{W}_i = 
    \begin{pmatrix}
      \bar W_{[1,1,i,1]} & \hdots & \bar W_{[w,h,i,1]} \\
        \vdots & \ddots & \vdots \\
      \bar W_{[1,1,i,n_{\text{out}}]} & \hdots & \bar W_{[w,h,i,n_{\text{out}}]} 
    \end{pmatrix}
\end{equation}
Thus, any convolutional layer can be converted to a depthwise separable convolution layer if we have $r_i = 1$ for all $i$. 
In such case a basis $\mathfrak{B}_i^1={D}_{[:,:,i,1]}$ of the rows of the matrix $\mathfrak{W}_i$ can trivially be retrieved by considering the first non-zero row and the whole layer can be transformed to a depthwise convolutional layer. In practice, we find that it happens often after hashing. Otherwise, for all $i$ such that $r_i \neq 1$ we denote ${(\mathfrak{B}_i^k)}$ a basis of the rows of $\mathfrak{W}_i$, with $k\in\llbracket 1; r_i \rrbracket$. 
Then, for every $j\in \llbracket 1; n_{\text{out}} \rrbracket$ there exists $\mu_i^k \in \mathbb{R}$ and $k\in\llbracket 1; r_y \rrbracket$ such that $\bar W_{[:,:,i,j]} = \mu_i^j \mathfrak{B}_i^k$.
Thus, in this case, each output channel $j$ gives rise to a number of basis kernels $(\mathfrak{B}_i^k)_{k=1 \dots r_i}$, and the corresponding subsequent point-wise convolutions with coefficients 
$\mu_i^j = {P}_{[1,1,i,j]}$, 
% $\mu_i^j = \lambda_i^j = {P}_{[1,1,i,j]}$, 
that depends of the rank of its concatenated flattened filter restriction matrices. 
Hence, we call this layer an uneven depthwise convolution with kernels defined as the basis kernels $\mathfrak{B}_i^k$. 
\newline\noindent
To sum it up, the proposed \ours method consists in three steps: an adaptive scalar hashing step (with hyperparameter $\tau$), followed by pruning \textit{via} similarity-based neuron merging (with hyperparameter $\alpha$), and an uneven depthwise separation step (summarized as algorithms in Section A.2 of the supplementary material).
As it will be shown in the experiments \ours significantly outperforms other data-free architecture compression methods and often matches data-driven methods.

\section{Experiments}\label{sec:experiments}
\subsection{Experimental setup}\label{sec:expsetup}
\paragraph{Datasets and baselines:} we evaluate our models on the two \textit{de facto} standard datasets for architecture compression, \textit{i.e.} CIFAR-10 (\citep{krizhevsky2009learning}, under the MIT License) and ImageNet (\citep{ImageNet_cvpr09}, under the BSD-3 license). We use the standard evaluation metric of removed parameters as well as removed FLOPs.
We apply our approach on ResNet (\citep{he2016deep}, ResNet 20-56-110 and 164 with respective number of parameters 270k, 852k, 1.7M and 2.6M, and accuracies $92.48$, $93.46$, $93.81$ and $94.54$ on CIFAR-10 and ResNet 50 with 25M parameters and $76.17$ accuracy on ImageNet) as well as Wide-ResNet \citep{zagoruyko2016wide}. Wide ResNet architectures are defined by their number of layers as well as their wideness multiplier: we evaluate on Wide ResNet 16-8 (with 11.0M and $95.2$ accuracy on CIFAR-10), 22-2 (with 1.5 and $94.1$ accuracy), 28 (28-2, 28-4, 28-8 and 28-10 with 1.9M, 7.4M, 29.8M and 36.5M parameters and $94.3$, $94.8$, $95.4$ and $95.8$ accuracies) and 40-4 (with 8.9M parameters and $95.0$ accuracy).
% \paragraph{Protocols:} in order to measure the efficiency of the proposed method relatively to the baseline networks, we report accuracy as a percentage of the baseline model accuracy or the accuracy drop. Similarly, for pruning, the percentage of remaining parameters w.r.t. the base model. To measure memory footprint reduction, we use the standard metric \cite{stock2019and} which consists in measuring the ratio between the zipped base model and the processed model.
\paragraph{Implementation details:}\label{sec:implem_detail}
The proposed adaptive hashing is implemented using Scikit-learn python library, with bandwidth $\Delta_l$ set as the median of the differences between consecutive weight values per layer $l$ and the contrast hyperparameter $\tau$ set to 0, by default. Merging and depthwise separation are implemented using Numpy: for relaxed merging, and unless stated otherwise, hyperparameter $\alpha$ is set to the highest value that fully preserves the model accuracy. We apply a per block strategy for setting the layer-wise $(\alpha^l)$ and a constant strategy for the $(\tau^l = \tau)$. Different strategy for setting the layer-wise $(\alpha^l)$ and $(\tau^l)$ are discussed in Section A.3 of the supplementary material.
We ran our experiments on a Intel(R) Core(TM) i7-7820X CPU. The proposed hashing method processing time depends on the model's size: ranging from 45 to 413 seconds for ResNets on CIFAR-10 and up to 21 hours for a ResNet-50 on ImageNet. This step could however be accelerated by a layer-wise parallelization. Pruning, on the other hand, is much faster, taking 5 seconds for ResNet-20 on CIFAR-10 and 15 minutes for a ResNet-50 on ImageNet.

\subsection{Hashing empirically induces vector and tensor-level redundancies}\label{sec:experiments_introspection}
\begin{figure}[t]
\begin{center}
\includegraphics[width = 0.8\linewidth]{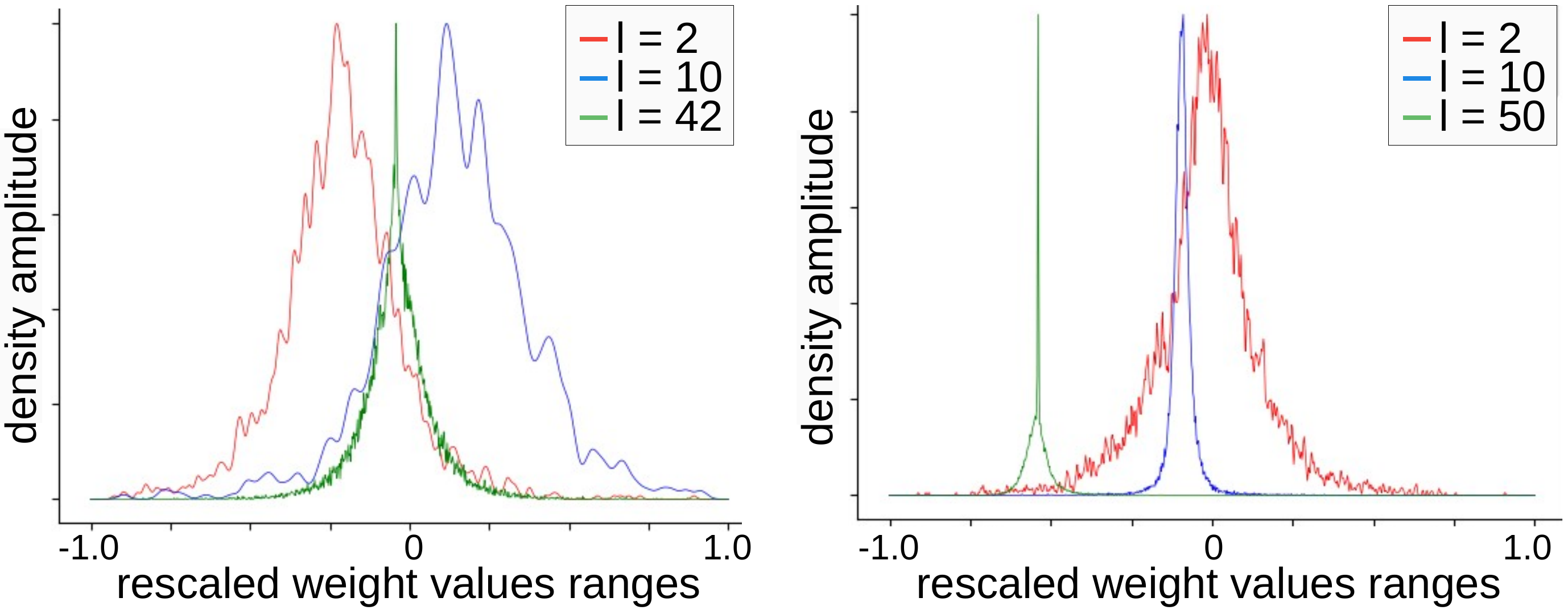}
\vskip -0.1in
\caption{
Weight distributions for several layers of a ResNet-56 on CIFAR-10 (left) and ImageNet (right) reveal that weights concentrate around multiple modes that can be captured \textit{via} hashing.
}
\label{fig:kde}
\end{center}
\vskip -0.1in
\end{figure}
First, Figure \ref{fig:kde} displays the weight distributions for several layers for ResNet-56 trained on CIFAR-10 and ResNet-50 on ImageNet. We observe that the weights concentrate around multiple dominant modes, with the number of modes and their relative proximity being variable from one layer to another: this motivates the design of the proposed adaptive hashing method, where each weight can be assigned to its corresponding mode, with optional collapse of adjacent modes.
\newline\noindent
%
% \textcolor{red}{a mettre dans le sup sous forme de tableau???}
%
In order to measure the impact of the change in the predictive function $f$ from the scalar hashing we considered the average error induced by hashing in logits $\mathbb{E}[|f(x) - \tilde f(x)|]$.
This error is compared to the distribution of the differences between the top1 and top2 logits.
An error smaller than the aforementioned difference implies that the modifications due to hashing can not change the prediction made by the DNN. For instance, we observe on ResNet 56 on CIFAR-10 an average error of 0.75 while the average difference between top 1 ad top 2 logits is $\approx 10.43$ in the baseline model with a standard deviation of 6.29. Furthermore, we observe similar values on other considered networks (for details see Appendix A.4), which empirically validates the efficiency of the hashing mechanism at preserving the accuracy.
\newline\noindent
\begin{figure*}[t]
\begin{center}
\includegraphics[width = \linewidth]{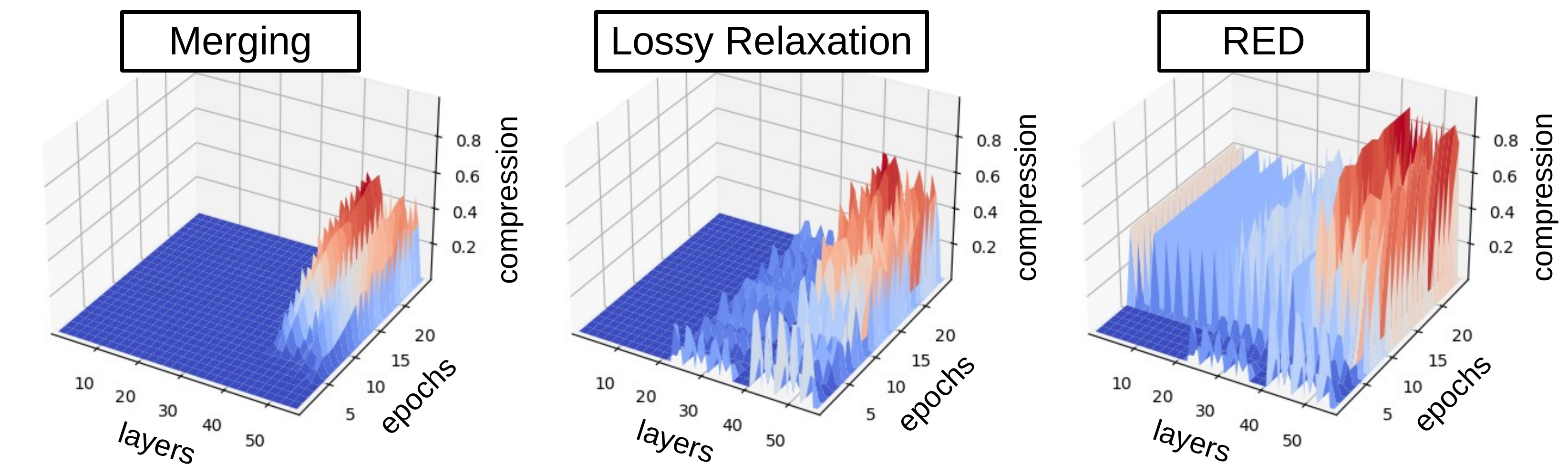}
% \vskip -0.05in
\caption{3D plot of the removed parameters ratio as a function of the layer's depth and the training epoch for ResNet-56 on ResNet 50 CIFAR-10. Regardless of the initialization schemes designed to avoid redundancies, the hashed networks present large numbers of vector and tensor redundancies, that can be removed with minimal impact on the accuracy.}
\label{fig:ResNet3D}
\end{center}
\vskip -0.2in
\end{figure*}\noindent
Figure \ref{fig:ResNet3D} also shows the evolution, during training and for each layer, of the removed parameter ratio (obtained with hashing and either merging, relaxed merging and \ours, which consists in applying relaxed merging plus uneven depthwise separation). Initially, for the first epochs, there are no vector redundancies (left and central plots) and few tensor redundancies (right plot), due to the initialization scheme (e.g. Glorot or He) specifically designed to avoid vector redundancies. However, we observe a rapid convergence towards high numbers of both vector and tensor redundant weight distributions after only 10 epochs. In fact, we show in Section A.5. of the supplementary material that we observe the same phenomenon even in the presence of training methods that aims at promoting diversity among neurons (e.g. dropout).
\newline\noindent
As such, after hashing, merging exact redundancies removes $0-50\%$ of the weights, depending on the layer, with more emphasis on the last layers. Relaxed merging (which looks after similar but not necessarily equal neurons) removes up to $60\%$ of the layers' weights without changing the network accuracy. Furthermore, we can remove a lot of parameters ($40-85\%$) with uneven depthwise separation among all the layers.

\begin{table*}[t]
\centering
\caption{Ablation results in terms of \% removed parameters compared to the base model.}
\begin{center}
% \begin{small}
\begin{tabular}{c |c c c c c c}
\hline
Hashing                    & \xmark & \xmark & \xmark & \cmark & \cmark & \cmark \\
Merge ($\alpha =0$)        & \cmark & \xmark & \cmark & \cmark & \cmark & \cmark \\
Merge ($\alpha =\alpha^*$) & \xmark & \cmark & \cmark & \xmark & \cmark & \cmark \\
Depthwise Separation       & \xmark & \xmark & \cmark & \xmark & \xmark & \cmark \\
\hline
ResNet 20  & 0.00 & 18.58 & 18.58 & 25.18 & 41.03 & \textbf{65.05}\\
ResNet 56  & 0.00 & 61.19 & 61.19 & 58.45 & 77.68 & \textbf{84.52}\\
ResNet 110 & 0.00 & 75.29 & 75.29 & 62.41 & 84.43 & \textbf{89.64}\\
ResNet 164 & 0.00 & 78.61 & 78.61 & 62.73 & 88.87 & \textbf{91.06}\\
\hline
Wide ResNet 16-8  & 0.00 & 31.08 & 31.08 & 19.67 & 38.67 & \textbf{51.92} \\
Wide ResNet 22-2  & 0.00 & 51.17 & 51.17 & 13.27 & 63.67 & \textbf{64.98} \\
Wide ResNet 28-2  & 0.00 & 49.19 & 49.19 & 11.46 & 61.20 & \textbf{64.19} \\  
Wide ResNet 28-4  & 0.00 & 41.79 & 41.79 & 20.99 & 51.99 & \textbf{56.07} \\
Wide ResNet 28-8  & 0.00 & 33.58 & 33.58 & 19.78 & 41.78 & \textbf{52.87} \\
Wide ResNet 28-10 & 0.00 & 47.25 & 47.25 & 25.59 & 58.79 & \textbf{60.80} \\
Wide ResNet 40-4  & 0.00 & 49.67 & 49.67 & 43.37 & 61.80 & \textbf{70.35} \\
\hline
\end{tabular}
\label{tab:cifar10_pruning}
% \end{small}
\end{center}
\end{table*}
\begin{figure*}[t]
\vskip -0.1in
\begin{center}
\centerline{\includegraphics[width=\linewidth]{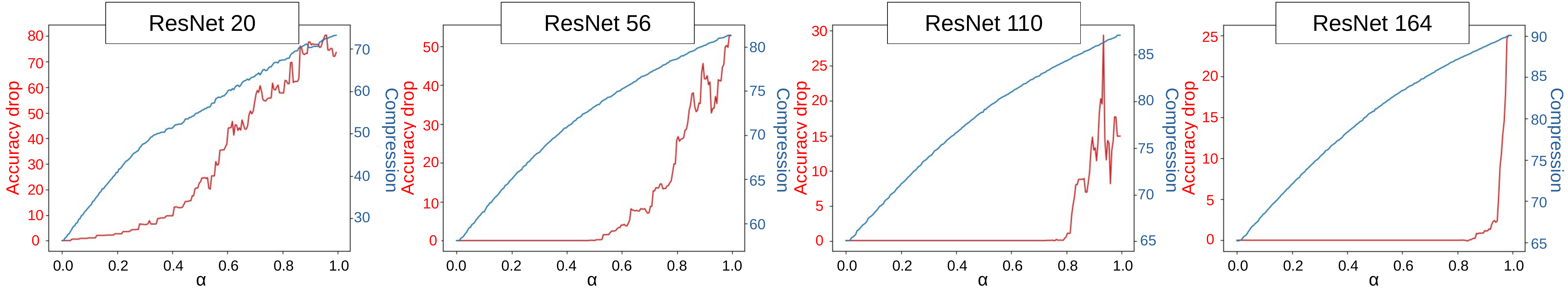}}
\caption{Accuracy drop ($\%$, red) and compression (in term of $\%$ removed parameters, blue) vs. values of $\alpha$ for ResNet 20, 56, 110 and 164 on CIFAR-10. We can remove high number of similar neurons without impacting the accuracy, particularly for deeper networks.}
\label{fig:LAC_ResNet}
\end{center}
\vskip -0.3in
\end{figure*}
\subsection{Ablation study}\label{sec:ablation}
% \paragraph{Memory footprint reduction:} Table~\ref{tab:cifar10_hashing} shows the memory footprint compression ratios obtained with each step of the proposed method. As depicted in the first column, hashing alone allows to compress the DNNs between $12\times$ and $36\times$ times, with heavier compression for deeper networks (ResNet 110 and 164 as well as Wide ResNet 28-8, 28-10 and 40-4). 
% Experiments on Wide ResNet demonstrate the robustness of our adaptive hashing algorithm to networks depth and width.
% Pruning \textit{via} merging redundant neurons also allows slightly higher compression rates. Furthermore, relaxed merging significantly increases the compression rate on certain networks (e.g. ResNet 110 and 164), by merging more neurons, still fully preserving the base model accuracy. Lastly, the complete \ours method, involving hashing, relaxed merging and uneven depthwise separation, enables superior compression rates ($21\times$ - $67\times$) with, to a certain extent, more compression for the deeper, wider networks. This shows the added value of each component of \ours in terms of memory footprint reduction.
% \paragraph{Runtime acceleration:}
Table~\ref{tab:cifar10_pruning} presents results in term of $\%$ removed parameters for each step in \ours for both ResNet and Wide ResNet architectures on CIFAR-10. First, we observe that without hashing, the $\%$ removed parameters is significantly lower for every model and pruning method or combination thereof.
For instance, on ResNet 20, Merge ($\alpha = 0$) with hashing (column 4) outperforms Merge ($\alpha = \alpha^*$) without hashing. Furthermore, it can be seen by comparing the first two columns that depthwise separation does not add much without hashing: therefore, we argue that scalar hashing is critical to introduce vector and tensor-level redundancies without accuracy loss.\newline\noindent
As such, hashing + merging ($\alpha=0$) already achieves good results without altering the predictive function (up to $62.7\%$ on deeper networks, e.g. ResNet-164). Furthermore, setting $\alpha=\alpha^*$ and using depthwise separation allows to reach higher removed parameters rates, e.g. up to $90\%$ on ResNet-110 and 164, without witnessing any accuracy drop.
The experiments on Wide-ResNets show that the thinner the network the higher the removed parameters rates, with Wide ResNet 40-4 and 22-2 having the highest removed parameters ratio and 16-8 and 28-8 having the lowest. This is natural since, from a combinatorial standpoint, assuming a similar prior, the chance to find redundancies (strict or approximate) reduces as the channel depth increases.
Last but not least, these results are very stable, as we observe standard deviations between $0.09-0.84$ for the different models over 10 runs of retraining and application of \ours.\newline
\begin{figure*}[t]
\begin{center}
 \includegraphics[width = \linewidth]{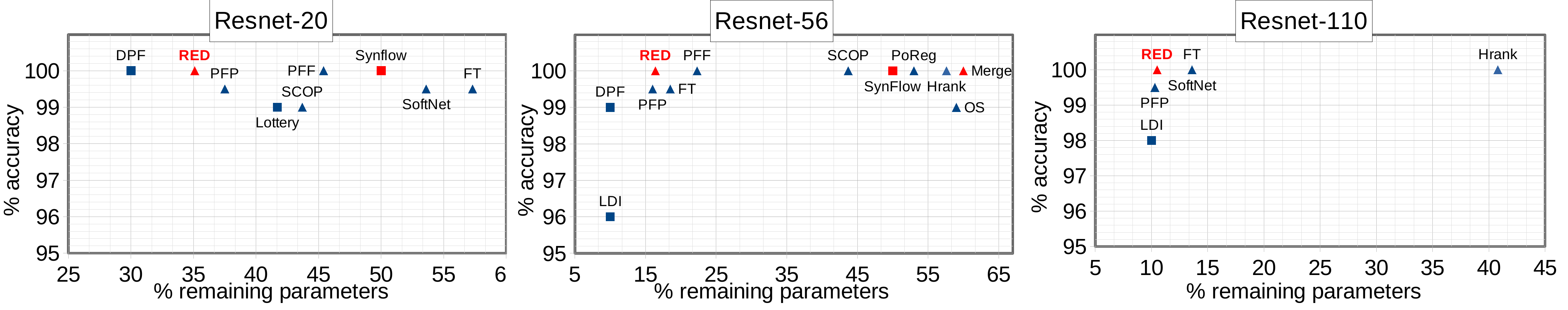}
\vskip -0.1in
\caption{Comparison between \ours and state-of-the-art methods on Cifar-10, in terms of $\%$ accuracy (measured as a percentage of the base model accuracy, the higher the better) and $\%$ remaining parameters (the lower the better). Each method is classified either as data-free (red) or data-driven (blue) and structured (triangle) or unstructured (rectangle). For each network, \ours performs significantly better than other data-free methods, and often as well as data-driven or unstructured methods.}
\label{fig:SOTA_pruning_Cifar10}
\end{center}
\vskip -0.2in
\end{figure*}
Figure \ref{fig:LAC_ResNet} shows the compression rate, and accuracy drop for relaxed merging as a function of parameter $\alpha$. For ResNet 20 we observe an initial low-slope phase for which merging neurons have a lower influence on the accuracy. For ResNet 56 and deeper, we observe an initial plateau where merging together a large number of similar neurons (up to $50\%$ and $80\%$ for ResNet 56 and 110-164, respectively) does not impact the accuracy at all.
This further illustrates that the hashed DNN (and particularly deeper networks) weights present structural redundancies, and that similarity-based analysis can dramatically increase their efficiency.
Note that the theoretical pruning factor of the merging and tensor decomposition steps are discussed in Section A.6 of the supplementary material. Also, we report the GPU and FLOPs performance of \ours in Section A.7. Finally, hashing also affects memory footprint as evaluated in Appendix A.8.

\subsection{Comparison with state-of-the-art approaches}\label{sec:experiments_SOTA}
\paragraph{Comparisons on CIFAR-10:}
Figure \ref{fig:SOTA_pruning_Cifar10} draws a comparison between \ours and state-of-the-art architecture compression methods, in terms of $\%$ remaining parameters and $\%$ accuracy w.r.t. the base model. For comparison purposes, we report results on the most popular architectures, \textit{i.e.} ResNet 20, 56 and 110. Each of these approaches is either classified as:
\begin{itemize}[noitemsep,topsep=0pt,parsep=0pt,partopsep=0pt]
\item Data-driven (blue): methods that rely on fine-tuning or retraining protocols requiring labelled data, such as LDI \citep{lee2019signal}, DPF \citep{lin2020dynamic}, PFP \citep{liebenwein2019provable}, FT \citep{li2016pruning}, SoftNet \citep{he2018soft}, Lottery \citep{frankle2018lottery}, PoReg \citep{zhuang2020neuron}, PFF \citep{meng2020pruning}, OS \citep{renda2020comparing} and SCOP \citep{tang2020scop}.
\item Data-free (red): methods that do not use data such as Merge \citep{kim2020neuron}, SynFlow \citep{tanaka2020pruning}, or methods that generate synthetic data for fine-tuning such as Dream \citep{yin2020dreaming}.
\end{itemize}
Another classification of these methods lies on the type of pruning: we distinguish unstructured methods (depicted by a square) from structured methods (triangle). 
Generally speaking, \ours performs significantly better than all the others state-of-the-art data-free methods: it outperforms its closest contenders, SynFlow \citep{tanaka2020pruning} on ResNet 20, as it removes $15\%$ more parameters without accuracy drop. 
When compared to the most similar method \citep{kim2020neuron}, the merging step alone (with $\alpha = 0$) already achieves $18\%$ higher removed parameters ratio on ResNet 56. Furthermore, the complete \ours approach significantly widens the gap and reach $44\%$ higher pruning ratio on ResNet 56. Furthermore, merging and \ours also allows higher removed parameters ratios on Wide ResNet 40-4 ($5.6\%$ and $30\%$, respectively), due to finer pairwise similarity modeling as well as handling structural redundancies through uneven depthwise separation.
\ours also outperforms the best structured, data-driven method, PFP \citep{liebenwein2019provable} by achieving similar compression ratios on the three networks without any accuracy drop. Last but not least, \oursvir, despite being a data-free approach, is competitive with recent data-driven and unstructured methods such as DPF \citep{lin2020dynamic}, Lottery \citep{frankle2018lottery} or LDI \citep{lee2019signal}.
\paragraph{Comparisons on ImageNet:}
\begin{table}[t]
    \centering
    \caption{Comparison with state-of-the-art methods in term of \% removed parameters/FLOPS and (top-1, top-5) accuracy on a ResNet-50 trained on ImageNet. Gray cells highlight data-free methods.}
\begin{center}
% \begin{small}
    \begin{tabular}{p{14mm}|c c c c}
        \hline 
        Model & \%removed  & \%removed & \hfil Top-1\% & \hfil Top-5\% \\
              & parameters &  FLOPs    &               & \\
        \hline
        Baseline                     & \hfil -   & \hfil -   & \hfil76.2 & \hfil92.9 \\
        \hline
        GAL-0.5 & 16.9 & 43.0 & 72.0 & 90.9 \\
        \cellcolor{Gray}Dream$_{20}$                      & \cellcolor{Gray}\hfil20.0 & \cellcolor{Gray}\hfil37.0 & \cellcolor{Gray}\hfil73.3 & \cellcolor{Gray}\hfil-    \\
        SSS-32 & 27.1 & 31.1 & 74.2 & 91.9\\
        $\text{Hrank}_{1.58}$                        & \hfil36.7 & \hfil43.7 & \hfil75.0 & \hfil92.3 \\
        \cellcolor{Gray}$\textbf{\ours}_{\tau = 0}$  & \cellcolor{Gray}\hfil39.6 & \cellcolor{Gray}\hfil42.7 & \cellcolor{Gray}\hfil76.1 & \cellcolor{Gray}\hfil92.9 \\
        GAL-1 & 42.5 & 61.4 & 69.9 & 89.8 \\
        SCOP                         & \hfil42.8 & \hfil45.3 & \hfil76.0 & \hfil92.8 \\
        \cellcolor{Gray}$\textbf{\ours}_{\tau = 5}$  & \cellcolor{Gray}\hfil42.1 & \cellcolor{Gray}\hfil44.5 & \cellcolor{Gray}\hfil75.3 & \cellcolor{Gray}\hfil92.1 \\
        $\text{Hrank}_{1.85}$                & \hfil46.0 & \hfil62.1 & \hfil72.0 & \hfil91.0 \\
        \cellcolor{Gray}$\textbf{\ours}_{\tau = 10}$ & \cellcolor{Gray}\hfil47.3 & \cellcolor{Gray}\hfil47.9 & \cellcolor{Gray}\hfil74.1 & \cellcolor{Gray}\hfil91.1 \\
        \cellcolor{Gray}Dream$_{50}$                  & \cellcolor{Gray}\hfil50.0 & \cellcolor{Gray}\hfil71.0 & \cellcolor{Gray}\hfil60.7 & \hfil \cellcolor{Gray}-   \\
        \cellcolor{Gray}$\textbf{\ours}_{\tau = 15}$ & \cellcolor{Gray}\hfil54.7 & \cellcolor{Gray}\hfil55.0 & \cellcolor{Gray}\hfil71.1 & \cellcolor{Gray}\hfil90.7 \\
        \cellcolor{Gray}$\textbf{\ours}_{\tau = 20}$ & \cellcolor{Gray}\hfil56.9 & \cellcolor{Gray}\hfil57.3 & \cellcolor{Gray}\hfil67.9 & \cellcolor{Gray}\hfil90.3 \\
        $\text{Hrank}_{2.64}$                 & \hfil67.7 & \hfil76.0 & \hfil69.1 & \hfil89.6 \\
        %ThiNet                       & \hfil66.0 & \hfil73.1 & \hfil68.4 & \hfil88.3 \\
        \hline
    \end{tabular}
    \label{tab:SOTA_pruning_ImageNet}
% \end{small}
\end{center}
\vskip -0.1in
\end{table}
Table~\ref{tab:SOTA_pruning_ImageNet} shows a comparison between \ours and other state-of-the-art structured pruning approaches on ImageNet. These methods allow to find different levels of compromise between the removed parameter ratio (as indicated by the $\%$ remaining parameters) and the accuracy drop. 
In order to find such trade-off with \oursvir, we vary the contrast hyperparameter $\tau$ in the hashing step: by merging together close modes of the weight distribution, we introduce more redundancies, allowing to further compress the network at the expense of accuracy. Although this initial hashing step may cause the network accuracy to drop, both redundant neuron merging and uneven depthwise separation induce negligible loss, as echoed by the previous experiments. Thus, \ours allows to remove between $39.6\%$ ($42.7\%$ FLOPs) of a ResNet-50 with only $0.1\%$ top-1 accuracy drop, and (with $\tau=20$) $56.9\%$ ($57.3\%$ FLOPs) with $8.3\%$ top-1 accuracy drop. Thus, \ours appears as a more efficient pruning algorithm than data-free Dream \citep{yin2020dreaming}, allowing to remove large numbers of parameters with minimal accuracy drop. Furthermore, \ours is once again very competitive with recent data-driven methods such as Hrank \citep{lin2020hrank}, SCOP \citep{tang2020scop}, SSS \citep{huang2018data} or GAL \citep{lin2019towards}.

% \subsection{Further Experiments}\label{sec:qualitative_analysis}
% Despite an initialization by random sampling (e.g. Glorot or He initialization), we observed, after a few epochs, DNN weights converge toward distributions with vector and tensor redundancies.
% We observe the same phenomenon in the presence of dropout \cite{srivastava2014dropout}, as showed in Section A.6 of the supplementary material.
% As such, merging exact redundancies removes $0-50\%$ of the weights, depending on the layer, with more emphasis on the last layers. Relaxed merging (which looks after similar but not necessarily equal neurons) removes up to $60\%$ of the layers' weights without changing the network accuracy. Note that the per block aspect of this central plot is a consequence of the $\alpha^l$ setting strategy which was tested alongside other strategies in Section A.2. Furthermore, we can remove a lot of parameters ($40-85\%$) with uneven depthwise separation among all the layers.

\section{Conclusion}\label{sec:conclusion}
In this paper, we proposed \ours, a novel data-free structured architecture compression method. First, \ours uses a novel adaptive scalar hashing of the weight distributions to introduce redundancies in DNNs under the form of vector redundancies as well as tensor redundancies. These redundancies can be exploited with similarity-based neuron merging, as well as a novel uneven depthwise separation scheme for convolutional layers, respectively. The whole approach is fully data-free. Furthermore, we demonstrated through thorough experiments involving several architectures and databases, that \ours significantly outperforms other data-free, structured pruning methods and often matches recent state-of-the-art data-driven pruning techniques, at a very minimal expanse in term of accuracy drop.

It shall nevertheless be empathized that \ours could theoretically be used in conjunction with other DNN compression techniques such as existing data-free quantization techniques. Furthermore, sparse pruning schemes, e.g. magnitude-based pruning methods could be considered to further reduce the computational runtime, given appropriate hardware. Last but not least, \ours could quite straightforwardly be applied to any existing off-the-shelf computer vision model where runtime optimization is a concern, such as, for instance, object detection or semantic segmentation architectures.

%empirically validate the use of hashing to introduce redundancies in weight distributions, as a pre-process for similarity-based DNN pruning. 
%The proposed \ours method, a data-free structured DNN pruning approach, tackles the runtime problem. 
%First, \ours involves a new data-free, adaptive hashing scheme based on per-layer weight value distributions density estimation which provides an efficient weight approximation and exhibits redundancies in DNN operations. 
%Second, we eliminate vector redundancies by merging together neurons based on a similarity measure.
%This process is improved by the scalar hashing pre-processing. 
%This method differs from the traditional magnitude-based absolute approaches. 
%Last but not least, we proposed a novel uneven depthwise separation of convolutional layers to factor redundant spatial components at the tensors level. 


\appendix
\section{Appendix}
\subsection{Commutativity}
\ours prunes a hashed neural network $\tilde f$ using two steps, a similarity-based pruning method (merge) and a tensor decomposition in an uneven depthwise separable convolution.
These two steps are commutative. 
Let's consider a layer $l$ with hashed weights $\tilde W^l$. 
Output dimensions of $\tilde W^l$ are merged if and only if they are considered similar (either identical or within the $\alpha^l\%$ most similar.
Let's note $i$ and $j$ the indices of two output dimensions that should be merged if merging was performed before the tensor decomposition.
Now if we perform the tensor decomposition first, we get two tensors $D^l$ and $P^l$ such that
\begin{equation}
    W^l_i = D^l \cdot P^l_i
\end{equation}
$W^l_i$ are the weights corresponding to the $i^\text{th}$ output and $\cdot$ is the channel-wise product.
As a direct consequence $\|W^l_i - W^l_j \| $ is among the $\alpha^l\%$ smallest distances if and only if  $\|P^l_i - P^l_j \| $ is also among the $\alpha^l\%$ smallest distances as long as the $D^l$ are normalized.
Thus, the pruning factor from merging is independent to the steps ordering.
In the case of $ \alpha = 0$ we have the same result for the uneven depthwise separable convolution.
This is simple to see as the ranks in the uneven depthwise are computed per input and the merging is done by output.

\subsection{Algorithm}
The proposed \ours method is summarized in algorithm \ref{alg:general}. Although our method is sequential, the two pruning steps can commute.
\begin{algorithm}[!h]
  \caption{RED method}
  \label{alg:general}
\begin{algorithmic}
  \STATE {\bfseries Input:} trained DNN $f$, hyper-parameters $\alpha$ and $\tau$
  \STATE $\tilde f\leftarrow$ Hashing ($f$, $\tau$) \hspace*{\fill} \textcolor{OliveGreen}{$\blacktriangleright$ Algorithm \ref{alg:hashing}}
  \STATE $\bar f\leftarrow$ Merging ($\tilde f$, $\alpha$) \hspace*{\fill} \textcolor{OliveGreen}{$\blacktriangleright$ Algorithm \ref{alg:merging}}
  \STATE $\bar f\leftarrow$ Depthwise\_Separation ($\bar f$) \hspace*{\fill} \textcolor{OliveGreen}{$\blacktriangleright$ Algorithm \ref{alg:depthwise}}
  \STATE return $\bar f$ 
\end{algorithmic}
\end{algorithm}
The first step of \ours is a data-free adaptive hashing step (Algorithm \ref{alg:hashing}) which transforms the trained neural network $f$ in a hashed version $\tilde f$. As seen in Section \ref{sec:experiments_introspection}, for certain layers, the KDE may have very close extremas, that can be fused depending of layer-wise hyperparameter $\tau^l$ (and global hyperparameter $\tau$), which defines the minimum contrast between two modes. However, note that setting $\tau^l =0$ still allows very efficient hashing.
\begin{algorithm}[!h]
  \caption{Hashing}
  \label{alg:hashing}
\begin{algorithmic}
  \STATE {\bfseries Input:} trained DNN $f$ with weights $(W^l)_{l \in \llbracket 1 ; L \rrbracket}$, hyper-parameters $(\tau^l)_{l \in \llbracket 1 ; L \rrbracket}$
  \STATE Initialize $\tilde f = f$ 
  \FOR{$l=1$ {\bfseries to} $L$}
  \STATE $d^l = \text{KDE}(W^l)$
    \STATE extract ${(m_k^l)}_{k \in K^-}$ and ${(M_k^l)}_{k \in K^+}$ from $d^l$
    \STATE ${(M_k^l)}_{k \in K^+}\leftarrow$ NMS $\left({(M_k^l)}_{k \in K^+}, \tau^l\right)$
    \FOR{$w \in W^l$}
    \STATE find $k$ such that $w \in [m_k^l; m_{k+1}^l[$
    \STATE $\tilde w \leftarrow M_k^l$
  \ENDFOR
  \ENDFOR
  \STATE return $\tilde f$ 
\end{algorithmic}
\end{algorithm}
The second step (Algorithm \ref{alg:merging}) consists in a similarity-based merging of neurons where the similarity is computed as the euclidean distance between the weight values corresponding to each neurons. The process also adequately updates the consecutive layers. For the sake of simplicity we consider a sequential model without skip connections in this implementation. The layer-wise hyperparameter $\alpha^l$ (and global hyperparameter $\alpha$) defines the proportion of non-identical neurons to remove after ranking the pairwise distance between them. In particular, for $\alpha = 0$, we only merge identical neurons.
\begin{algorithm}[!h]
  \caption{Merging Redundancies}
  \label{alg:merging}
\begin{algorithmic}
  \STATE {\bfseries Input:} hashed DNN $\tilde f$, hyper-parameters $(\alpha^l)_{l \in \llbracket 1 ; L \rrbracket}$
  \STATE Initialize $\bar f = \tilde f$ with $(\bar W^l)_{l \in \llbracket 1 ; L \rrbracket} \leftarrow (\tilde W^l)_{l \in \llbracket 1 ; L \rrbracket}$
    \FOR{$l=1$ {\bfseries to} $L-1$}
    \STATE $D \leftarrow$ matrix of $l^2$ distances between all neurons
    \STATE $d \leftarrow \alpha^l$ percentile of $D$ \hspace*{\fill}\textcolor{OliveGreen}{$\blacktriangleright$ $d$ is the threshold distance}
    \STATE $D_{i,j} \leftarrow$ $1_{D_{i,j} \geq d \text{ or } i =j}$ \hspace*{\fill} \textcolor{OliveGreen}{$\blacktriangleright$ $D$ is a graph of neurons connected by similarity}
    \STATE $M \leftarrow$ connected components from $D$
  \STATE $\bar W^{l}_{\text{new}} = []$
    \FOR{comp $\in M$} 
    \STATE $\bar W^{l}_{\text{new}}\text{.append}\left(\frac{1}{\left|\text{comp}\right|} \sum_{j\in\text{comp}} \bar W^{l}_{[...,j]}\right)$ \hspace*{\fill} \textcolor{OliveGreen}{$\blacktriangleright$ merge per connected component} 
  \ENDFOR
  \STATE $\bar W^l \leftarrow \bar W^{l}_{\text{new}}$
  \STATE $\bar W^{l+1}_{\text{new}} = []$ \hspace*{\fill} \textcolor{OliveGreen}{$\blacktriangleright$ We still have to update the layer $l+1$}
  \FOR{comp $\in M$}
  \STATE $\bar W^{l+1}_{\text{new}}\text{.append}\left(\sum_{i \in \text{comp}} \bar W^{l+1}_{[i,...]}\right)$
  \ENDFOR
  \STATE $\bar W^{l+1} \leftarrow \bar W^{l+1}_{\text{new}}$
  \ENDFOR
  \STATE return $\bar f$ 
\end{algorithmic}
\end{algorithm}
The final step, only relevant to CNNs, (Algorithm \ref{alg:depthwise}), checks if convolutional layers can be converted in depthwise separable ones based on the criterion we introduced. Note that we didn't describe the situation where ranks are not all equal to 1 as this case only changes the depthwise implementation and basis extraction.
\begin{algorithm}[!h]
  \caption{Depthwise Separation}
  \label{alg:depthwise}
\begin{algorithmic}
  \STATE {\bfseries Input:} merged DNN $\bar f$ with weights $(\bar W^l)_{l \in \llbracket 1 ; L \rrbracket}$
  \FOR{$\bar f^l$ convolutional layer of shape $w,h,n_{\text{in}},n_{\text{out}}$}
  \FOR{$i=1$ {\bfseries to} $n_{\text{in}}$}
  \STATE $r_i \leftarrow \text{rank}
    \begin{pmatrix}
      \bar W_{[1,1,i,1]} & \hdots & \bar W_{[w,h,i,1]} \\
        \vdots & \ddots & \vdots \\
      \bar W_{[1,1,i,n_{\text{out}}]} & \hdots & \bar W_{[w,h,i,n_{\text{out}}]} 
    \end{pmatrix}$
    \STATE $D_{[:,:,i,1]} \leftarrow \bar W_{[...,i,j]}$ for $j$ such that $W_{[...,i,j]}\neq 0$
    \STATE $P_{[1,1,i,j]} \leftarrow \bar W_{[x,y,i,j]} / D_{[x,y,i,j]}$ such that $D_{[x,y,i,j]} \neq 0$ 
  \ENDFOR
  \STATE $f_d \leftarrow$ Depthwise conv layer of weight $D$
  \STATE $f_p \leftarrow$ Pointwise conv layer of weight $P$
  \STATE $\bar f^l \leftarrow f_p \circ f_d$ 
  \ENDFOR
  \STATE return $\bar f$ 
\end{algorithmic}
\end{algorithm}

\subsection{Hyper-parameters Application Strategy}
In addition to a trained neural network, our method takes two hyper-parameters $\alpha$ and $\tau$ as inputs.
The hyperparameter $\tau$ defines the average value of the per layers $\tau^l$ contrast hyperparameters of the adaptive hashing step. The modes within range $\tau^l\%$ of the total range of the distributions are collapsed to the maximum value among them. The hyperparameter $\alpha$ defines the average value of the per layers $\alpha^l$ proportion of non-identical neurons to merge for each layer. For each of these hyper-parameters we compare different strategies to allocate values to each individual $(\alpha^l)$ from $\alpha$ and $(\tau^l)$ from $\tau$. For $(\alpha^l)$ we tested the following strategies:
\begin{itemize}
    \item per \textit{block} strategy: we group $\alpha^l$ values per $1/3$ of the network layers such that:
    \begin{equation}
    \begin{cases}
        \alpha^l = \max\{2\alpha -1,0\} & \text{if } l \in \llbracket 0 ; L/3 \llbracket \\
        \alpha^l = \alpha & \text{if } l \in \llbracket L/3 ; 2L/3 \rrbracket \\
        \alpha^l = \min\{2\alpha,1\} & \text{if } l \in \rrbracket 2L/3 ; L \rrbracket \\
    \end{cases}
    \end{equation}
    \item \textit{constant} strategy: $\forall l \in \llbracket 1 ; L \rrbracket$, $\alpha^l$
    \item \textit{linear ascending} strategy: $\alpha^l$ $\forall l \in \llbracket 1 ; L \rrbracket$, $\alpha^l = \alpha l / L$
    \item \textit{linear descending} strategy: $\forall l \in \llbracket 1 ; L \rrbracket$, $\alpha^l = \alpha (L-l) / L$
\end{itemize}
\begin{table}[!t]
\centering
\caption{Comparison between different strategies for $\alpha^l$ in terms of pruning factor for ResNet 56 on CIFAR-10, with constant $\tau=0$.}
\vspace{0.25cm}
\begin{tabular}{l|c}
\hline
Strategy & \% removed parameters \\
\hline
linear descending & 77.90\\
\hline
constant  & 78.69\\
\hline
linear ascending & 80.35\\
\hline
block  & \textbf{84.52} \\
\hline
\end{tabular}
\label{tab:1}
\end{table}

Table \ref{tab:1} draws a comparison between these different strategies in term of pruning ratio, with $\alpha$ set to the minimal value that does not bring any accuracy loss. We found the \textit{linear descending} strategy, despite allowing to remove nearly $80\%$ of the network parameters, to be the least performing one, wollowed by the \textit{constant} strategy. The \textit{linear ascending} strategy is significantly better, validating the general idea that shallower layers contain more redundant information. Following this idea, the best performing strategy \textit{block} allows to remove more than $4\%$ extra parameters: hence, we keep this per-block strategy in others experiments.

\begin{figure}[!t]
\begin{center}
% \vskip -0.1in
\includegraphics[width = 0.6\linewidth]{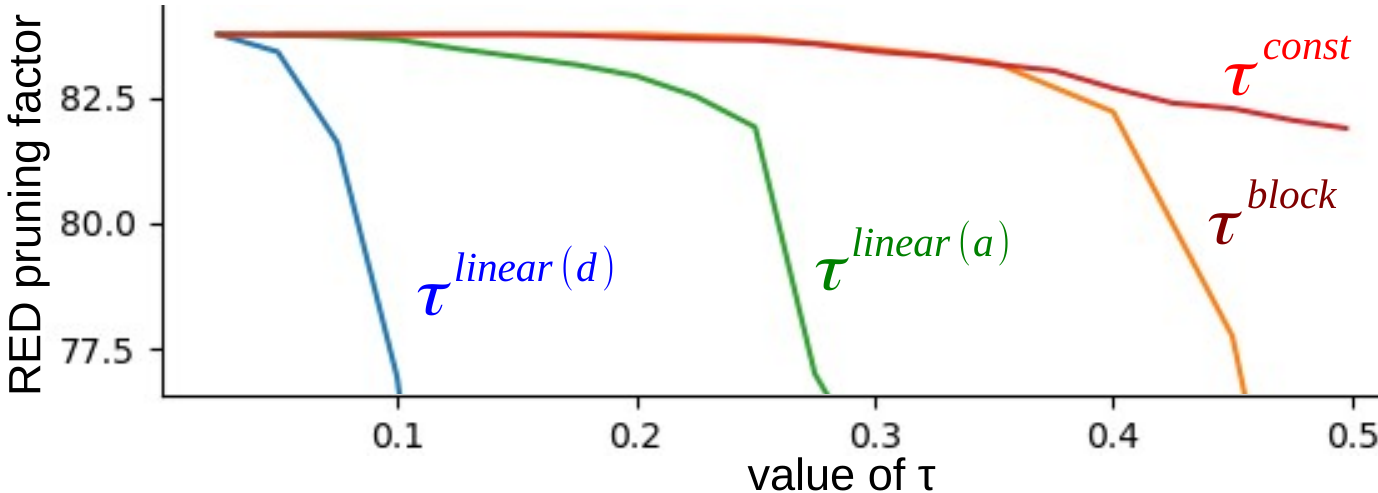}
\vskip -0.1in
\caption{Evolution of the performance of RED in term of \% removed parameters as a function of $\tau$ for the four tested strategies. The \textit{constant} strategy provides the best results.}
\label{fig:tau}
\end{center}
\vskip -0.1in
\end{figure}

We evaluated the same strategies, \textit{block}, \textit{constant}, \textit{linear ascending} and \textit{linear descending} for the hashing contrast hyperparameters $(\tau^l)$. To evaluate each strategy for $\tau$ we measure the \% of parameters removed with \ours while keeping the accuracy constant: thus, the higher the pruning factor (with equal accuracy), the better the method. The results are showcased on Figure \ref{fig:tau} for values of $\tau$ ranging from 0 to $0.5$. We see that the \textit{linear descending} and \textit{linear ascending} strategies are the least performing, followed by the \textit{block} strategy. The \textit{constant} strategy is the best performing, thus we keep this constant setting in our experiments.

\subsection{Impact of Hashing}
\begin{table}[!th]
\centering
\caption{Comparison between the modification induced by hashing (second column) and the confidence of the baseline model (third column), \textit{i.e.} the difference between the highest and second highest logits.}
\begin{tabular}{l|c|c}
\hline
model & $\mathbb{E}[|f(x) - \tilde f(x)|]$ & $\mathbb{E}[|\text{top}_1(f(x)) - \text{top}_2(f(x))|]$ \\
\hline
\hline
ResNet 20  & 2.90 & 9.56 \\
ResNet 56  & 0.75 & 10.43 \\
ResNet 110 & 1.48 & 11.18 \\
ResNet 164 & 2.69 & 11.28 \\
\hline
Wide ResNet 28-10 & 1.24 & 10.95 \\
Wide ResNet 40-4  & 0.63 & 10.88 \\
\end{tabular}
\label{tab:hash_OReOS}
\end{table}
Following the study from Section \ref{sec:experiments_introspection}, we want to empirically validate that hashing a DNN $f$ doesn't change the predictions.
To do so we compare the error between the hashed DNN $\tilde f$ and the baseline model $f$ to the difference between the top1 and top2 logits in the original prediction from $f$.
This is based on the fact that if the error is lower than the difference between the two logits with the highest responses then the highest logit will remain unchanged.
We already provided values for ResNet 56 in the main paper.
In Table \ref{tab:hash_OReOS}, we provide values for more networks we benchmarked in pruning.
As Stated in the main paper, the modifications from hashing are significantly smaller than the difference between the two highest logits of the baseline model.
This is a beneficial consequence of the over-confidence of modern DNNs.
This empirical observation validates the accuracy preservation after hashing and suggests that similar results could be achieved on other over-confident DNNs.

\subsection{Robustness to Dropout}
\begin{figure}[!t]
\begin{center}
\includegraphics[width = 0.65\linewidth]{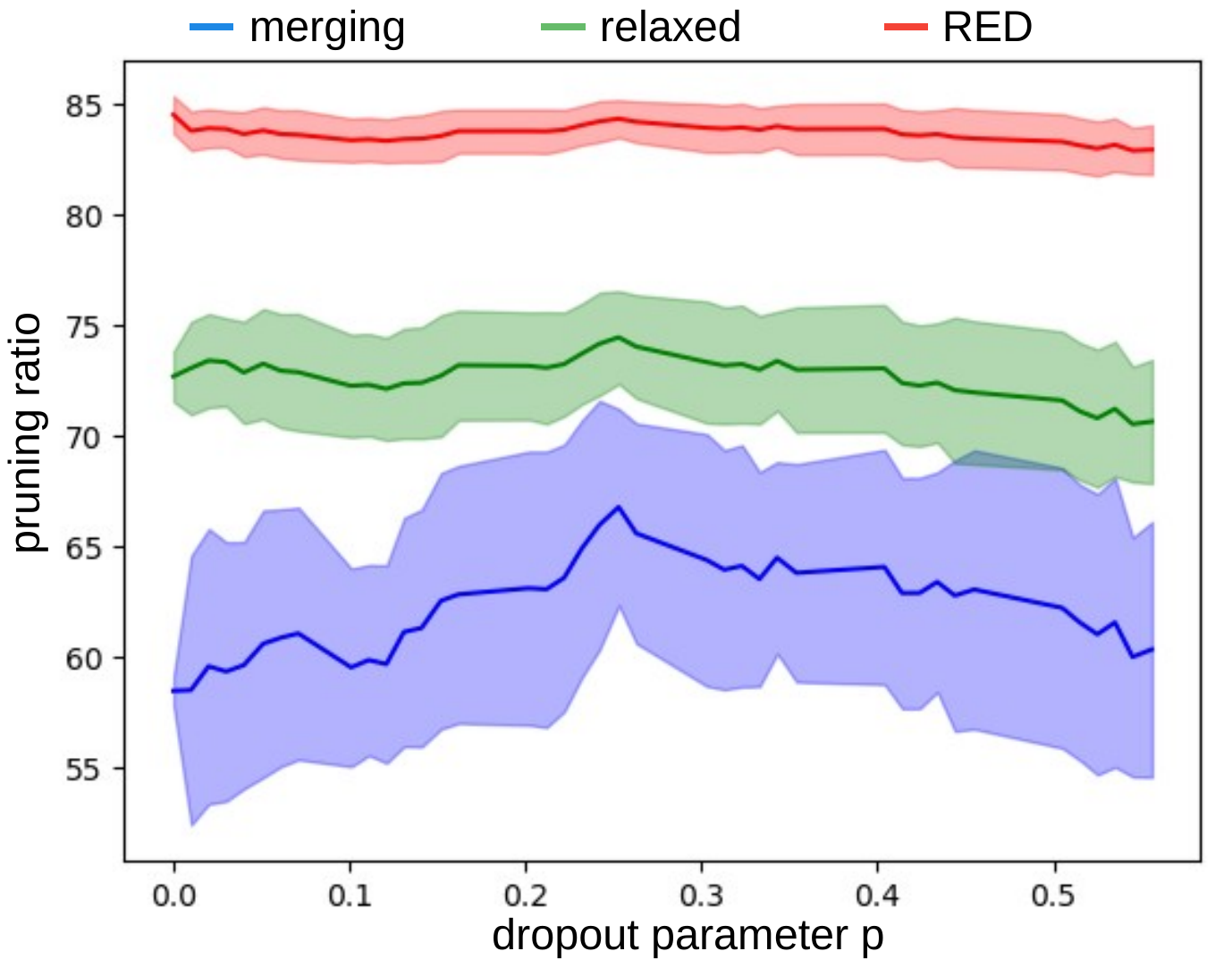}
\caption{Graphs of the pruning factor resulting from the steps of \ours as functions of the dropout parameter $p$. \ours appears to be robust to dropout.}
\label{fig:dropout}
\end{center}
\vskip -0.2in
\end{figure}

Despite initialization methods used to avoid redundancies within the layers weights, other training or regularization methods exists with the purpose of exploiting available weights, thus potentially reducing redundancies, such as Dropout \citep{srivastava2014dropout}. Dropout is a vastly used deep learning technique that aims at avoiding overfitting by preventing co-adaptation between neurons. In practice, this is done by randomly dropping neurons at train time with a probability $p$. In order to evaluate the robustness of \ours to different values of $p$ (modulating the intensity of the dropout), we retrained a ResNet-56 on CIFAR-10 with different values for $p$ on the last layer and then applied \ours on these trained networks. Figure \ref{fig:dropout} shows the \% removed parameters (average and standard deviation over 10 experiments) for each step of RED as a function of the dropout parameter $p$. 
% We observe that the average performance of the proposed method (red curve) are stable across the tested values for $p$. 
First, surprisingly, we observe that Dropout causes an important rise of the average performance for the merging with $\alpha =0$ (blue curve), with a significant variance. This is likely due to the fact the Dropout affects the last layers where most redundancies are found as seen in Figure \ref{fig:ResNet3D} of the main paper. However, this effect is mitigated by the merging relaxation before completely vanishing after the uneven depthwise separation step. 
% This can be explained by the rigidity of the pruning factor induced by the depthwise separation detailed in Section \ref{sec:A3}. 
Furthermore we empirically observe that the ranks $r_i$ still converge to $1$ in presence of Dropout. 
% As empirically shown in Section 4.2 networks rapidly learn weight kernel which can be transformed in depthwise separable convolutions, which are only affected by dropout in the point-wise layer thus suggesting that ranks shouldn't be modified due to Dropout.
Based on these observations we can assess that, overall, the proposed method \ours appears to be very robust to dropout. 

\subsection{Expected Pruning Factor}\label{sec:A3}
The pruning factor can be estimated from the number of unique neurons per layer and the hyper-parameter $\alpha$. Let's consider a convolutional layer $l$ with shape $w\times h \times n_{\text{in}}\times n_{\text{out}}$ and an input $\alpha$ for the merging step and the corresponding $\alpha^l \in [0;1[$ which indicates the proportion of unique neurons that shall be merged. Then, the pruned weight tensor will have a shape $w\times h \times n_{\text{in}}\times \lfloor \gamma^l (1-\alpha^l) n_{\text{out}} \rceil$, where $\lfloor \cdot \rceil$ is the rounding operation and $\gamma^l$ is the proportion of unique neurons. The pruning ratio $r_{\text{merge}}^l$ at the end of this step is
\begin{equation}
    r_{\text{merge}}^l = \gamma^l (1- \alpha^l)
\end{equation}
After the merging step we apply our depthwise separation technique to further compress the network, with the resulting pruning ratio $r_{\text{RED}}^l:$
 \begin{equation}
 \begin{cases}
    r_{\text{RED}}^l = \frac{w h + r_{\text{merge}}^ln_{\text{out}}}{w h n_{\text{out}}}\quad\quad\quad\text{ if }\forall i \in\llbracket 1; n_in\rrbracket,  r_i = 1\\
    r_{\text{RED}}^l = \frac{w h + r_{\text{merge}}^ln_{\text{out}}}{w h n_{\text{out}}} \frac{\sum_{i=1}^{n_{\text{in}}} r_i}{n_{\text{in}}}\\
 \end{cases}
 \end{equation}
 where the $r_i$ are the ranks of the matrix obtained from the per input channel flattened weights, concatenated along the output channel.
 
\subsection{FLOPs and Inference-time}
\begin{table*}[!t]
\centering
\caption{Percentage of FLOPs removed for different models on Cifar10.}
\begin{center}
\begin{small}
\begin{tabular}{c |c c c c c c}
\hline
 & \multicolumn{6}{c}{\% removed FLOPS} \\
\hline
Hashing                    & \xmark & \xmark & \xmark & \cmark & \cmark & \cmark \\
Merge ($\alpha =0$)        & \cmark & \xmark & \cmark & \cmark & \cmark & \cmark \\
Merge ($\alpha =\alpha^*$) & \xmark & \cmark & \cmark & \xmark & \cmark & \cmark \\
Depthwise Separation       & \xmark & \xmark & \xmark & \xmark & \xmark & \cmark \\
% \hline
% \multirow{3}{*}{Model} & \multicolumn{6}{c}{\% removed FLOPs} \\
% \cline{2-7}
%  & \multicolumn{2}{c}{merging ($\alpha = 0$)} & \multicolumn{2}{c}{relaxed ($\alpha = \alpha^*$)} & \multicolumn{2}{c}{\oursvir} \\
% \cline{2-7}
% & no hashing & hashing & no hashing & hashing & no hashing & hashing\\
\hline
ResNet 20  & 0 & 17.33 & 17.33 & 27.13 & 42.90 & \textbf{63.00} \\
ResNet 56  & 0 & 60.22 & 60.22 & 58.57 & 77.61 & \textbf{81.72} \\
ResNet 110 & 0 & 75.10 & 75.10 & 63.01 & 84.45 & \textbf{87.98} \\
ResNet 164 & 0 & 76.89 & 76.89 & 63.26 & 86.90 & \textbf{91.43} \\
\hline
Wide ResNet 16-8  & 0 & 30.78 & 30.78 & 20.07 & 39.20 & \textbf{52.04} \\
Wide ResNet 22-2  & 0 & 51.34 & 51.34 & 13.50 & 63.82 & \textbf{65.09} \\
Wide ResNet 28-2  & 0 & 50.87 & 50.87 & 11.16 & 60.91 & \textbf{64.10} \\  
Wide ResNet 28-4  & 0 & 41.72 & 41.72 & 21.07 & 51.56 & \textbf{55.99} \\
Wide ResNet 28-8  & 0 & 33.28 & 33.28 & 19.80 & 41.62 & \textbf{53.17} \\
Wide ResNet 28-10 & 0 & 47.39 & 47.39 & 25.54 & 58.69 & \textbf{59.78} \\
Wide ResNet 40-4  & 0 & 49.51 & 49.51 & 43.41 & 61.99 & \textbf{70.09} \\
\end{tabular}
\label{tab:cifar10_pruning2}
\end{small}
\end{center}
\vskip -0.2in
\end{table*}
\begin{table*}[!t]
\centering
\caption{Runtime gain as a percentage of the removed inference time for different Cifar10 models on different hardware.}
\begin{center}
\begin{small}
\begin{tabular}{c | c | c | c | c| c }
\hline
device & batch size & ResNet 20 & ResNet 56 & ResNet 110 & ResNet 164 \\
\hline
RTX 3090 (GPU) & 256 & 62\% & 75\% & 85\% & 89\% \\
% RTX 6000 (GPU) & 64 & & & & \\
% i7-7820X (CPU) & 64 & & & & \\
Intel m3 (CPU) & 32 & 87\% & 88\% & 88\% & 89\% \\
\hline
\end{tabular}
\label{tab:cifar10_time}
\end{small}
\end{center}
\vskip -0.2in
\end{table*}
In Section \ref{sec:ablation} we evaluated \ours using the standard metric of proportion of removed parameters. Another classic metric is the proportion of removed FLOPs which we provide here in Table \ref{tab:cifar10_pruning2}. We observe that the results are similar to the values from Table \ref{tab:cifar10_pruning}. This is a consequence of the proposed pruning protocol which removes parameters in a structured way and removes more parameters on large convolutions (deep layers) as can be seen in Figure \ref{fig:ResNet3D}. Another important cause is the pruning of $3\times 3$ convolutional layers which represent a large proportion of both parameters and FLOPs. Note that FLOPs removal are already provided for ImageNet in Table \ref{tab:SOTA_pruning_ImageNet}.\newline\noindent
Another intuitive metric for pruning evaluation would be to compare runtime on CPU/GPU. However this metric presents many flaws, among which dependencies on the batch size, hardware and use of inference engines. Nonetheless we report the runtime gains over different hardware in Table \ref{tab:cifar10_time}. We measured the proportion of reduced computation time (i.e. the higher the better). We observe that on very small CPU (e.g. Intel m3) many operations are slowing the inference down, thus networks compression vastly impacts inference speed, although on ResNet 164 some operations remain bottleneck as the speed-up doesn't grow much with the pruning ratio. On the other hand, for large GPU (e.g. RTX 3090) we observe that the inference time reduction if strongly correlated to the pruning ratio and FLOPs removal. Overall \ours enables a very effective speed-up of DNN inference from $60\%$ to $90\%$ while preserving the accuracy.

\subsection{Memory footprint Reduction}\label{sec:A6}
\begin{table*}[!t]
\centering
\caption{Ablation results in terms of memory footprint reduction (ratio between zipped base and processed models).}
\begin{center}
\begin{small}
\begin{tabular}{c|c|c|c|c}
\hline
\multirow{2}{*}{Model} & \multicolumn{4}{c|}{\cellcolor{Gray}zipped model memory ratio} \\
\cline{2-5}
 & hashing & $\alpha = 0$ & $\alpha = \alpha^*$ & \ours  \\
\hline
ResNet 20  & 12.36  & 12.86 & 12.95 & \textbf{21.69} \\
ResNet 56  & 23.29  & 25.71 & 27.76 & \textbf{41.34} \\
ResNet 110 & 35.74  & 38.45 & 43.04 & \textbf{58.33} \\
ResNet 164 & 25.40  & 25.40 & 52.92 & \textbf{66.84} \\
\hline
Wide ResNet 16-8  & 17.05 & 21.95 & 21.95 & \textbf{41.26} \\
Wide ResNet 22-2  & 12.07 & 12.93 & 12.98 & \textbf{30.17} \\
Wide ResNet 28-2  & 12.00 & 12.67 & 12.91 & \textbf{28.50} \\  
Wide ResNet 28-4  & 15.15 & 17.88 & 17.88 & \textbf{33.11} \\
Wide ResNet 28-8  & 18.90 & 25.93 & 25.93 & \textbf{50.76} \\
Wide ResNet 28-10 & 20.32 & 29.34 & 29.34 & \textbf{47.00} \\
Wide ResNet 40-4  & 17.62 & 26.24 & 26.24 & \textbf{47.63} \\
\hline
\end{tabular}
\label{tab:cifar10_hashing}
\end{small}
\end{center}
\end{table*}\noindent
In this work we focus on pruning, nonetheless we also studied the consequence on the memory footprint of DNNs.
To measure this impact we consider the ratio of the size of the processed networked zipped over the size of the original network also zipped.
The empirical results are listed in Table \ref{tab:cifar10_hashing}.
We observe that the hashing step alone has large influence on the memory footprint dividing it by $12$ on already small networks (e.g. ResNet 20 and Wide ResNet 28-2) and up to $35$ on larger networks (e.g. ResNet 110).
The memory footprint is further reduced by pruning and tensor decomposition.
Reaching $20$ times reduction on ResNet 20 and up to $67$ times on ResNet 164.
Note that the zipping process aplies Huffman coding which depends on the distribution of the values to zip. 
For instance less values closer to a uniform distribution will less compressed that a larger list with a more peaky distribution.
This is the case for some networks as their pruned version are not significantly smaller on disk once zipped compared to their hashed version.

{
\small
\bibliography{main.bib}
\bibliographystyle{icml2021}
}

\end{document}